\documentclass{article}
\usepackage{latexsym}
\usepackage{amsmath}
\usepackage{amsthm}
\usepackage{amsfonts}
\usepackage{url}
\usepackage{amssymb}
\IfFileExists{mytheo.sty}
  {\usepackage{mytheo}}{\usepackage{../../latex_files/mytheo}}
\usepackage{graphicx}
\usepackage{color}

\usepackage{ifpdf}
\ifpdf    %
\usepackage{hyperref}
\else    %
\usepackage[hypertex]{hyperref}
\fi

\def\balig{\begin{IEEEeqnarray*}{L}}
\def\ealig{\end{IEEEeqnarray*}}
\def\balign{\begin{IEEEeqnarray}{L}}
\def\ealign{\end{IEEEeqnarray}}

\usepackage{hyperref}
\usepackage{url}
\usepackage{mathrsfs}
\usepackage{latexsym}
\usepackage{amsmath}
\usepackage{amsfonts}
\usepackage{amssymb}
\usepackage{amsthm}
\usepackage{enumitem}

\usepackage{times}
\usepackage{graphicx} %
\usepackage{algorithm}
\usepackage{algorithmicx,algpseudocode}
\usepackage{mathrsfs}
\usepackage{latexsym}
\usepackage{tikz}
\usetikzlibrary{positioning}
\usepackage{epstopdf}
\usepackage{hyperref}

\usepackage{graphicx} %
\usepackage{float}

\newcommand{\err}{\operatorname{err}}
\newcommand{\serr}{\widehat{\err}}

\renewcommand{\paragraph}[1]{\noindent {\bf#1}}

\newcommand{\dist}{\rho} %

\newcommand{\marg}{\gamma}

\mathchardef\mhyphen="2D

\newcommand{\ddim}{\operatorname{ddim}}

\newcommand{\argmin}{\mathop{\mathrm{argmin}}}

\newcommand{\inv}{^{-1}} %

\newcommand{\X}{\calX}

\usepackage{dsfont}
\newcommand{\chr}{\mathds{1}}
\newcommand{\pred}[1]{\chr_{\left\{ #1 \right\}}}

\newcommand{\diam}{\operatorname{diam}}

\renewcommand{\P}{\mathbb{P}}

\newcommand{\ben}{\begin{enumerate}}
\newcommand{\een}{\end{enumerate}}
\newcommand{\bit}{\begin{itemize}}
\newcommand{\eit}{\end{itemize}}

\def\clap#1{\hbox to 0pt{\hss#1\hss}}

\def\pskip{\vspace{-0.2cm}}

\newcommand{\calX}{\mathcal{X}}

\newcommand{\R}{\mathbb{R}}
\newcommand{\N}{\mathbb{N}}

\newcommand{\beq}{\begin{eqnarray*}}
\newcommand{\eeq}{\end{eqnarray*}}
\newcommand{\beqn}{\begin{eqnarray}}
\newcommand{\eeqn}{\end{eqnarray}}
\newcommand{\paren}[1]{\left( #1 \right)}

\newcommand{\tlprn}[1]{\left\{ #1 \right\}}
\newcommand{\set}[1]{\tlprn{#1}}

\newcommand{\ceil}[1]{\ensuremath{\left\lceil#1\right\rceil}}

\newcommand{\ds}{\displaystyle}

\newcommand{\oo}[1]{\frac{1}{#1}}
\def\eps{\varepsilon}

\newtheorem{theorem}{Theorem}

\newtheorem{lemma}[theorem]{Lemma}

\newtheorem{corollary}{Corollary}
\renewcommand{\bepf}{\begin{proof}}
\renewcommand{\enpf}{\end{proof}}

\newcommand{\citep}[1]{\cite{#1}}
\newcommand{\citet}[1]{\cite{#1}}

\begin{document}
\title{Near-optimal sample compression for nearest neighbors
}

\author{Lee-Ad Gottlieb\thanks{
L. Gottlieb is with the 
Department of Computer Science at Ariel University
(email: leead@ariel.ac.il).
His work was supported in part by 
the Israel Science Foundation (grant No. 755/15)
},
Aryeh Kontorovich\thanks{
A. Kontorovich is with the
Department of Computer Science at Ben-Gurion University of the Negev
(email: karyeh@cs.bgu.ac.il).
His work was supported in part by 
the Israel Science Foundation (grants No. 1141/12 and 755/15)
and 
a Yahoo Faculty award.
} and
Pinhas Nisnevitch\thanks{
  P. Nisnevitch is an M.Sc. student
  at the Computer Science department of Tel-Aviv University
  and
  may be reached at (email: pinhasn@mail.tau.ac.il).
}
}

\maketitle

\begin{abstract}
We present the first sample compression algorithm for nearest neighbors
with non-trivial performance guarantees. We complement these guarantees 
by demonstrating almost matching hardness lower bounds, which show
that our performance bound is nearly optimal.
Our result yields new insight into margin-based nearest neighbor
classification in metric spaces and allows us to significantly sharpen and simplify
existing bounds. 
Some encouraging empirical results are also presented.
\end{abstract}

\section{Introduction}
The nearest neighbor classifier for non-parametric classification is perhaps the most
intuitive learning algorithm. 
It is apparently the earliest, having been introduced
by Fix and Hodges in 1951 (technical report reprinted in \citet{FH1989}). In this model, 
the learner observes a sample $S$ of 
labeled 
points $(X,Y) = (X_i,Y_i)_{i\in[n]}$, 
where $X_i$ is a point in some metric space $\X$ and $Y_i \in \{-1,1\}$ 
is its label.
Being a metric space, $\X$ is equipped with a distance function 
$\dist: \X\times\X \rightarrow \R$. Given a new unlabeled point $x\in\X$ 
to be classified, 
$x$ is assigned the same label as its nearest neighbor in $S$, which is
$\argmin_{Y_i \in \set{-1,1}} \dist(x,X_i)$. 
Under mild regularity assumptions, 
the nearest neighbor classifier's expected error is asymptotically bounded by 
twice the Bayesian error, 
when the sample size tends to infinity
\citep{CoverHart67}.\footnote{A Bayes-consistent modification of the $1$-NN classifier was recently
proposed in \cite{kon-weiss-2015}.}
These results have inspired a vast
body of research 
on proximity-based classification
(see \citet{MR2097754,shwartz2014understanding} for extensive background and 
\cite{DBLP:journals/corr/ChaudhuriD14} for a recent refinement of classic results).
More recently, 
strong margin-dependent generalization bounds were obtained in
\citet{DBLP:journals/jmlr/LuxburgB04},
where the margin is the minimum distance between
opposite labeled points in $S$.

In addition to provable generalization bounds, nearest neighbor (NN) classification 
enjoys several other
advantages. 
These include simple evaluation on new data, immediate extension to multiclass 
labels, and minimal structural assumptions --- it does not assume a Hilbertian or even a
Banach space. 
However, the naive NN approach also has disadvantages. 
In particular,
it requires
storing the entire sample,
which may be memory-intensive. 
Further, information-theoretic considerations show that 
exact NN evaluation requires
$\Theta(|S|)$ time in high-dimensional metric spaces \citep{KL04}
(and possibly Euclidean space as well \citep{Cla-94})
---
a phenomenon known as the algorithmic
{\em curse of dimensionality}. 
Lastly, the NN classifier has infinite VC-dimension 
\citep{shwartz2014understanding}, implying 
that it tends to overfit the data. 
This last problem can be mitigated by taking the majority 
vote among $k>1$ nearest neighbors 
\citep{MR1311980,MR1635410,shwartz2014understanding}, or 
by deleting some sample points so as to attain a larger margin 
\citep{DBLP:journals/tit/GottliebKK14}.

Shortcomings in the NN classifier led Hart \citet{DBLP:journals/tit/Hart68}
to pose the problem of sample compression.
Indeed, significant compression 
of the sample 
has the potential to simultaneously
 address the issues of memory usage, NN search time,
and overfitting. 
Hart considered the minimum Consistent Subset
problem --- elsewhere called the Nearest Neighbor Condensing problem ---  
which seeks to identify
a minimal subset $S^* \subset S$ that is {\em consistent} with $S$, 
in the sense that 
the nearest neighbor in $S^*$ of every $x \in S$ possesses the same label as $x$.
This problem is known to be NP-hard \citep{Wil-91,Zuk-10}, and Hart 
provided a heuristic 
with runtime $O(n^3)$.
The runtime of this heuristic was recently improved
by \citet{DBLP:conf/icml/Angiulli05}
to $O(n^2)$,
but neither paper gave approximation guarantees.

The Nearest Neighbor Condensing problem has been the subject
of extensive research since its introduction \citep{gates72,ritter75,wilson00}. 
Yet surprisingly,
there are no known approximation algorithms for it --- all previous results on this 
problem are heuristics that lack any non-trivial approximation guarantees. 
Conversely,
no strong hardness-of-approximation results for this problem are known, 
which indicates a gap in the current state of knowledge.

\paragraph{Main results.}
Our contribution aims at closing the existing gap in solutions to
the Nearest Neighbor Condensing problem.
We present a simple near-optimal approximation algorithm for this problem, 
where our only structural assumption is that the points lie in some metric space.
Define the {\em scaled margin} $\gamma < 1$ 
of a sample $S$ as the ratio of the minimum distance between opposite labeled 
points in $S$ to the diameter of $S$.
Our algorithm produces a consistent set 
$S' \subset S$ of size $\lceil 1/\gamma \rceil^{\ddim(S)+1}$ 
(Theorem~\ref{thm:alg}), where $\ddim(S)$
is the doubling dimension of the space $S$.
This result can significantly speed up evaluation on test points, and also
yields sharper and simpler generalization bounds than were previously known
(Theorem~\ref{thm:gen}).

To establish optimality, we complement the approximation result
with an almost matching 
hardness-of-approximation lower-bound. 
Using a reduction from the Label Cover problem,
we show that the Nearest Neighbor Condensing problem 
is NP-hard to approximate within 
factor $2^{(\ddim(S) \log (1/\gamma))^{1-o(1)}}$ 
(where $\ddim(S)$ or $\gamma$ is a function of $n$, see Theorem~\ref{thm:hard}).
Note that the above upper-bound is an absolute size guarantee, 
and stronger than an approximation guarantee.

Additionally, we present a simple heuristic 
to be applied in conjunction with the algorithm of Theorem~\ref{thm:alg},
that achieves further sample compression.
The empirical performances of both our algorithm and heuristic 
seem encouraging (see Section~\ref{sec:exper}).

\paragraph{Related work.}
A well-studied problem related to the Nearest Neighbor Condensing problem is that
of extracting a small set of simple conjunctions consistent with much of the 
sample, 
introduced by \citet{DBLP:journals/cacm/Valiant84} and shown 
by \citet{Hau-88} to be equivalent to minimum Set Cover
(see \citet{DBLP:journals/ml/LavioletteMSS10,DBLP:journals/jmlr/MarchandS02} 
for further extensions).
This problem is monotone in the sense that adding a 
conjunction to the solution set can only increase the 
sample accuracy of the solution. In contrast,
in our problem the addition of a point of $S$ to $S^*$ can cause $S^*$ to be 
inconsistent
--- and this distinction is critical to the hardness of our problem. 

Removal of points from the sample can also yield lower dimensionality, which itself
implies faster nearest neighbor evaluation and better generalization
bounds. For metric spaces, \citet{GK-13} and \citet{DBLP:conf/alt/GottliebKK13} 
gave algorithms for
dimensionality reduction via point removal (irrespective of margin size).

The use of doubling dimension as a tool to characterize metric
learning has appeared several times in the literature, initially by
\citet{BKL06} in the context of nearest neighbor classification, and
then in \citet{LL06} and \citet{Bshouty2009323}.  A series of papers
by Gottlieb, Kontorovich and Krauthgamer investigate doubling spaces
for classification
\citep{DBLP:journals/tit/GottliebKK14},
regression
\citep{GottliebKK13-simbad+IEEE}, and dimension reduction
\citep{DBLP:conf/alt/GottliebKK13}.

\paragraph{$k$-nearest neighbor.}
A natural question is whether the Nearest Neighbor
Condensing problem of \citet{DBLP:journals/tit/Hart68}
has a direct analogue when the
$1$-nearest neighbor rule is replaced by a $(k>1)$-nearest
neighbor -- that is, when the label of a point is determined by 
the majority vote among its $k$ nearest neighbors. 
A simple argument shows that the analogy breaks down.
Indeed, a minimal requirement for the 
condensing problem to be meaningful is that the full (uncondensed) set 
$S$ is feasible, i.e.\ consistent with itself. Yet even for 
$k=3$ there exist self-inconsistent sets. Take for example the set
$S$ consisting of two positive points at $(0,1)$ and $(0,-1)$ and two 
negative points at $(1,0)$ and $(-1,0)$. Then the $3$-nearest 
neighbor rule misclassifies every point in $S$, hence $S$
itself is inconsistent.

\paragraph{Paper outline.}
This paper is organized as follows.
In Section~\ref{sec:alg}, we present our algorithm and prove its performance bound,
as well as the reduction implying its near optimality (Theorem~\ref{thm:hard}).
We then highlight the implications of this algorithm for learning in Section~\ref{sec:learn}.
In Section~\ref{sec:exper} we describe a heuristic which refines our algorithm, and
present empirical results.

\subsection{Preliminaries}

\paragraph{Metric spaces.}
A {\em metric} $\dist$ on a set $\X$ is a positive symmetric function
satisfying the triangle inequality $\dist(x,y)\leq \dist(x,z)+\dist(z,y)$; together the two comprise 
the metric space $(\X,\dist)$.
The diameter of a set $A\subseteq\X$,
is defined by $\diam(A)=\sup_{x,y\in A}\dist(x,y)$.
Throughout this paper we will 
assume that 
$\diam(S)=1$;
this can always be achieved by scaling.

\paragraph{Doubling dimension.}
For a metric $(\X,\dist)$, let
$\lambda$
be the smallest value such that every ball in $\X$ of radius $r$ (for any $r$)
can be covered by $\lambda$ balls of radius $\frac{r}{2}$.
The {\em doubling dimension} of $\X$ is $\ddim(\X)=\log_2\lambda$.
A metric is {\em doubling}
when its doubling dimension is bounded. Note that while a low Euclidean
dimension implies a low doubling dimension (Euclidean metrics of dimension
$d$ have doubling dimension $O(d)$ \citep{DBLP:conf/focs/GuptaKL03}), low doubling
dimension is strictly more general than low Euclidean dimension.
The following packing property can be demonstrated via a repetitive application of
the doubling property:
For set $S$ with doubling dimension $\ddim(\X)$ and $\diam(S)\leq\beta$, 
if the minimum interpoint distance in $S$ is at least $\alpha < \beta$ then
\beqn
\label{eq:packing}
|S| \le \ceil{\beta/\alpha}^{\ddim(\X) +1}
\eeqn
(see for example \citet{KL04}). The above bound is tight up to constant factors
in the exponent,
meaning there exist sets of size $(\beta/\alpha)^{\Omega(\ddim(\X))}$.

\paragraph{Nearest Neighbor Condensing.} 
Formally, we define the Nearest Neighbor Condensing (NNC) problem as follows: We 
are given a set $S = S_- \cup S_+$ of points, and distance metric 
$\dist: S \times S \rightarrow \R$. We must 
compute a minimal cardinality subset $S' \subset S$ with the property that for
any $p \in S$, the nearest neighbor of $p$ in $S'$ comes from the same subset
$\{S_+,S_-\}$ as does $p$. If $p$ has multiple exact nearest neighbors in $S'$, 
then they must all be of the same subset.

\paragraph{Label Cover.}
The Label Cover problem was first introduced by \citet{ABSS-93}
in a seminal paper on the hardness of computation. 
Several formulations of this problem have appeared the literature, 
and we give the description forwarded by \citet{DS-04}:
The input is a bipartite graph $G=(U,V,E)$, with two sets of labels: 
$A$ for $U$ and $B$ for $V$. 
For each edge $(u,v) \in E$ (where $u \in U$, $v \in V$), 
we are given a relation
$\Pi_{u,v} \subset A \times B$ consisting of admissible 
label pairs for that edge. 
A {\em labeling} $(f,g)$
is a pair of functions 
$f:U \rightarrow 2^{A}$
and 
$g:V \rightarrow 2^{B} \backslash \{\emptyset\}$
assigning a set of labels to each vertex.
A labeling {\em covers} an edge $(u,v)$ 
if for every label 
$b \in g(v)$
there is some label
$a \in f(u)$
such that 
$(a,b) \in \Pi_{u,v}$. 
The goal is to find a labeling that covers all edges, 
and which minimizes the sum of the number of
labels assigned to each $u \in U$, that is
$\sum_{u \in U} |f(u)|$.
It was shown in
\citet{DS-04} that it is NP-hard to
approximate Label Cover to within a factor $2^{(\log n)^{1-o(1)}}$, where $n$
is the total size of the input.

In this paper, we make the trivial assumption that each vertex has some edge incident to it.
For ease of presentation, we will make the additional assumption that 
the label relations associate unique labels to each vertex. More formally, if
label pair $(a,b)$ is admissible for edge $(u,v)$
and $(a,b')$ is admissible for $(u',v')$,
then $u$ and $u'$ must be the same vertex.
Similarly if
$(a,b)$ is admissible for edge $(u,v)$
and $(a',b)$ is admissible for $(u',v')$,
then $v$ and $v'$ must be the same vertex.
This amounts to a naming convention, and has no effect on the problem instance.

\paragraph{Learning.}
We work in the {\em agnostic} learning model \citep{mohri-book2012,shwartz2014understanding}.
The learner receives $n$ labeled examples $(X_i,Y_i)\in\X\times\set{-1,1}$ drawn
iid 
according to some unknown
probability distribution $\P$.
Associated to any {\em hypothesis} $h:\X\to\set{-1,1}$ is its {\em empirical error}
$\serr(h)=n\inv \sum_{i\in[n]}\pred{h(X_i)\neq Y_i}$
and {\em generalization error}
$
\err(h)=\P(h(X)\neq Y).
$

\section{Near-optimal approximation algorithm}
\label{sec:alg}
In this section, we describe a simple approximation algorithm for the 
Nearest Neighbor Condensing problem. In Section \ref{sec:hard}
we provide almost tight hardness-of-approximation bounds.
We have the following theorem:

\begin{theorem}\label{thm:alg}
Given a point set $S$ and its scaled margin $\gamma < 1$, there exists an algorithm that in time 
$$\min \{n^2, 2^{O(\ddim(S))} n \log \lceil 1/ \gamma \rceil \}$$
computes a consistent set $S' \subset S$ of size at most
$\lceil 1/\gamma \rceil^{\ddim(S)+1}.$
\end{theorem}

Recall that an $\eps$-net of point set $S$ is a subset $S_\eps \subset S$ with two properties:
\begin{itemize}
\item[(i)] {\em Packing.}
The minimum interpoint distance in $S_\eps$ is at least $\eps$.
\item[(ii)] {\em Covering.} Every point $p \in S$ has a nearest neighbor in $S_\eps$
{\em strictly} within distance $\eps$. 
\end{itemize}
We make the following observation:
Since the margin of the point set is $\gamma$, a $\gamma$-net of $S$ is consistent
with $S$. That is, every point $p \in S$ has a neighbor in $S_{\gamma}$ strictly
within distance $\gamma$, and since the margin of $S$ is $\gamma$, this
neighbor must be of the same label set as $p$. By the packing property of doubling 
spaces (Equation \ref{eq:packing}),
the size of $S_{\gamma}$ is at most $\lceil 1/\gamma \rceil^{\ddim(S)+1}$. 
The solution returned by our algorithm is
$S_{\gamma}$, and satisfies the guarantees claimed in Theorem \ref{thm:alg}.

It remains only to compute the net $S_{\gamma}$. A brute-force greedy algorithm 
can accomplish this in time $O(n^2)$: For every point 
$p \in S$, we add $p$ to $S_{\gamma}$ if the distance from $p$ to all points 
currently in $S_{\gamma}$ is $\gamma$ or greater, $\dist(p,S_\gamma) \ge \gamma$.
See Algorithm \ref{alg:bf}.

\pskip

\begin{algorithm}
\caption{Brute-force net construction}
\label{alg:bf}
\begin{algorithmic}[1]
\Require $S$
\State $S_\gamma \leftarrow$ arbitrary point of $S$
\ForAll{$p \in S$}
    \If{$\dist(p,S_\gamma) \ge \gamma$}
        \State $S_\gamma = S_\gamma \cup \{p\}$
    \EndIf
\EndFor
\end{algorithmic}
\end{algorithm}

\pskip

The construction time can be improved by building a {\em net hierarchy},
similar to the one employed by \citet{KL04}, in total time
$2^{O(\ddim(S))} n \log(1/ \gamma)$.
(See also \citet{BKL06,HM06,CG06}.)
A hierarchy consists of nets
$S_{2^i}$ for $i = 1,0,\ldots,\lfloor \log \gamma \rfloor$,
where $S_{2^i} \subset S_{2^{i-1}}$ for all $i>\lfloor \log \gamma \rfloor$.
Further each point $p \in S$ is {\em covered} by at least one point in 
$S_{2^i}$, meaning there exists $q \in S_{2^i}$ 
satisfying 
$\dist(p,q) < 2^i$,
and hence
$\dist(p,S_{2^i}) < 2^i$.
Finally, we say that two points $p,q \in S_{2^i}$ are 
{\em neighbors} in net $S_{2^i}$ 
if $\dist(p,q) < 4 \cdot 2^i$.
Note that if $p,q$ are neighbors, and $p',q' \in S_{2^{i+1}}$ are
the respective covering points of $p,q$, then
$p',q'$ are necessarily neighbors in net $S_{2^{i+1}}$:
$\dist(p',q') 
\le \dist(p',p) + \dist(p,q) + \dist(q,q')
< 2^{i+1} + 4 \cdot 2^i + 2^{i+1}
= 4 \cdot 2^{i+1}$.

The net $S_{2^1} = S_2$ consists of a single arbitrary point of $S$.
Having constructed $S_{2^i}$, it is an easy matter to
construct $S_{2^{i-1}}$: 
First, since we require 
$S_{2^{i-1}} \supset S_{2^i}$, 
we will initialize 
$S_{2^{i-1}} = S_{2^{i}} $.
Now for each $p \in {S - S_{2^i}}$, we must determine whether 
$\dist(p,S_{2^{i-1}}) \ge 2^{i-1}$, and if so add
$p$ to $S_{2^{i-1}}$.
Crucially, we need not compare $p$ to all points of $S_{2^{i-1}}$: 
If there exists $q \in S_{2^{i-1}}$, satisfying
$\dist(p,q) < 2^{i-1}$, then $p,q$ are neighbors in net $S_{2^{i-1}}$,
and so their respective covering points $p',q' \in S_{2^i}$ are
neighbors in net $S_{2^i}$. 
Let set $T$ include only the neighbors of $p'$ and the points of 
$S_{2^{i-1}}$ covered by these neighbors; it suffices to compute whether 
$\dist(q,T) \ge 2^{i-1}$.
The points of $T$ have minimum distance $2^{i-1}$ and are
all contained in a ball of radius $4 \cdot 2^i+2^{i-1}$ centered
at $q'$, so by the packing property (Equation \ref{eq:packing})
$|T| = 2^{O(\ddim(S))}$.
It follows that the above query $\dist(q,T)$ can be answered in
time $2^{O(\ddim(S))}$. For each point in $S$ we execute
$O(\log \lceil 1/\gamma \rceil)$ queries, for a total runtime of
$2^{O(\ddim(S))} n \log \lceil 1/\gamma \rceil$.
The above procedure is illustrated in
Algorithm \ref{alg:fast} in the Appendix.

\subsection{Hardness of approximation of NNC}\label{sec:hard}

In this section, we prove almost matching hardness results for the NNC problem.

\begin{theorem}\label{thm:hard}
Given a set $S$ of labeled points with scaled margin $\gamma$, 
it is NP-hard to approximate the
solution to the Nearest Neighbor Condensing problem on $S$ to within a factor
$2^{(\ddim(S) \log (1/\gamma) )^{1-o(1)}}$,
where $\ddim(S)$ or $\gamma$ is a function of $n$.
\end{theorem}

\pskip

To simplify the proof, we introduce an easier
version of NNC called {\em Weighted} Nearest Neighbor Condensing (WNNC). 
In this problem, the input is augmented with a function assigning weight 
to each point
of $S$, and the goal is to find a subset $S' \subset S$ of minimum
{\em total weight}. We will reduce Label Cover to WNNC and then reduce 
WNNC to NNC, all
while preserving hardness of approximation. The theorem will follow from 
the hardness of Label Cover \citep{DS-04}.

\begin{figure*}[ht]
\vskip 1cm
\begin{center}
\centerline{\includegraphics[width=1\columnwidth]{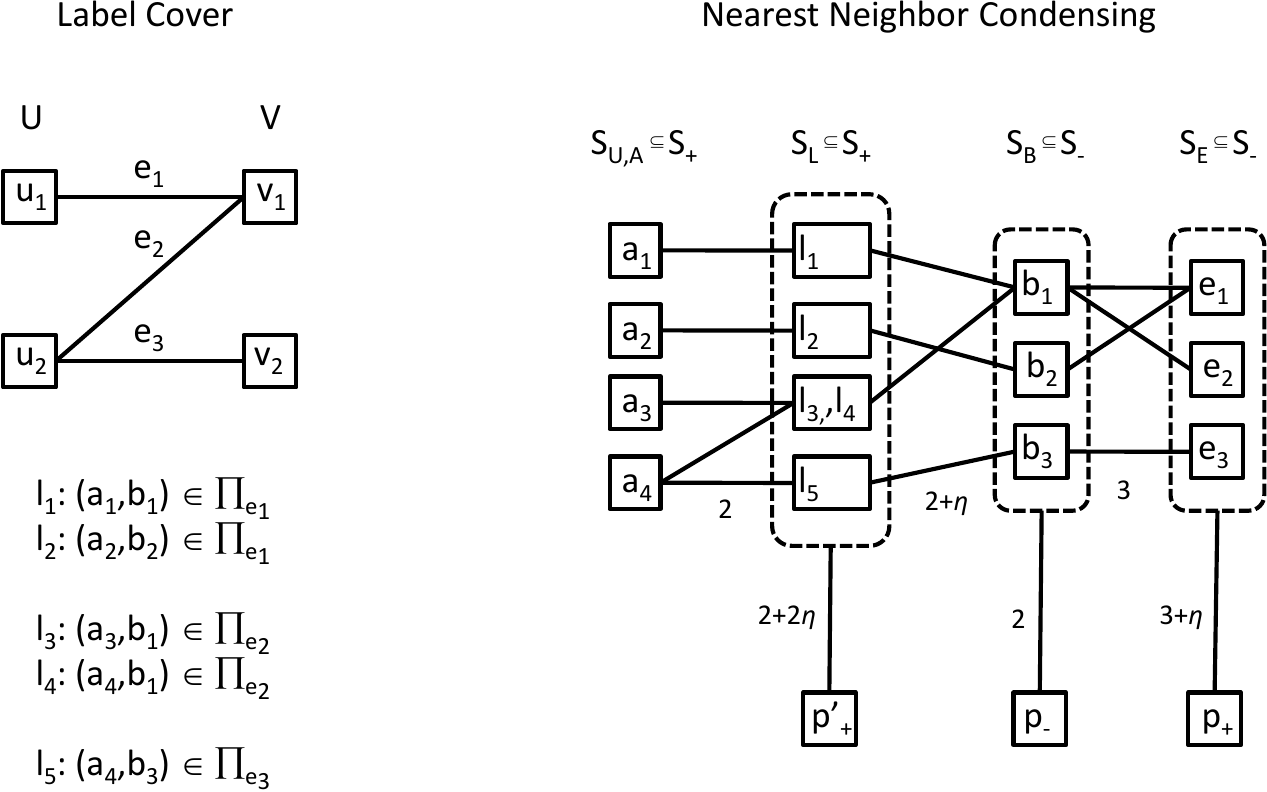}}
\caption{
Reduction from Label Cover to Nearest Neighbor Condensing.}
\label{fig:NNE3}
\end{center}
\end{figure*}

\paragraph{First reduction.}
Given a Label Cover instance of size
$m = |U|+|V|+|A|+|B|+|E|+\sum_{e\in E}|\Pi_e|$,
fix an infinitesimally small constant $\eta$.
We create an instance of WNNC as 
follows (see Figure~\ref{fig:NNE3}).

\begin{enumerate}[leftmargin=.5cm]
\item
We introduce set $S_E \subset S_-$ representing edges in $E$:
For each edge $e \in E$, create point $p_e$ of weight $\infty$.
We also create a point $p_+ \in S_+$ of weight 0, and
the distance from $p_+$ to each $p_e \in S_E$ is $3+\eta$.

\item
We introduce set $S_B \subset S_-$ representing labels in $B$:
For each label $b \in B$, create point $p_b$ of weight 0.
If $b$ is found in an admissible label for edge $e$, 
then the distance from $p_b$ to $p_e$ is 3.
We also create a point $p_- \in S_-$ of weight 0, 
at distance 2 from all points in $S_B$.

\item
We introduce set $S_L \subset S_+$ representing labels in $\Pi_e$.
For each edge $e$ and label $b \in B$ that is part of an admissible pair for $e$, 
we create point $p_{e,b} \subset S_L$ of weight $\infty$.
This point represents all label pairs in $\Pi_e$ that contain $b$.
$p_{e,b}$ is at distance $2+\eta$ from $p_{b}$.
We also create a point $p'_+ \in  S_+$ of weight 0, at distance 
$2+2\eta$ from all points in $S_L$.

\item
We introduce set $S_A \subset S_+$ representing labels in $A$:
For each label $a \in A$, create point $p_a$ of weight $1$.
If $a$ is part of an admissible pair for any label of 
$p_{e,b}$, then then the distance 
from $p_{a}$ to $p_{e,b} \in S_L$ is 2.
\end{enumerate}

The points of each set $S_E$, $S_B$, $S_L$ and $S_A$ are packed into
respective balls of diameter 1. Let $g$ be the minimum inter-point distance 
within each set. Since each set has cardinality less than $m$, we have
$m = (1/g)^{O(\ddim(S))}$,
or equivalently
$g = m^{-O(1/\ddim(S))}$.
All interpoint distances not yet specified are set to their maximum possible value.
The diameter of the resulting set is constant, as is its scaled margin.

We claim that a solution of WNNC on the constructed instance implies a solution
to the Label Cover Instance with the same cost. Briefly, points in $S_E \subset S_-$ have
infinite cost, so they cannot be included in the solution. Since they are close
to $p_+$, they must be
covered by points in $S_B \subset S_-$ -- this corresponds to choosing labels in $B$ for
each point in $V$, where the labels must be admissible for the edges incident
to $V$. Similarly, the points in $S_L \subset S_+$ have infinite cost, so if they 
they are close to points in $S_B \subset S_-$, then they must be
covered by some points in $S_A \subset S_+$. This corresponds to choosing labels in $A$ for
points in $U$, where the labels must complement the labels in $B$ previously
chosen, and complete the cover of all edges. More formally:

\begin{enumerate}[leftmargin=.5cm]
\item
$p_+$ must appear in any solution: The nearest neighbors of $p_+$
are the negative points of $S_E$, so if $p_+$ is not included
the nearest neighbor of set $S_E$ is necessarily the nearest neighbor
of $p_+$, which is not consistent.
\item
Points in $S_E$ have infinite weight, so no points of 
$S_E$ appear in the solution. All points of $S_E$ are at distance
exactly $3+\eta$ from $p_+$, hence each point of $S_E$ must be covered
by some point of $S_B$ to which it is connected -- other points in
$S_B$ are farther than $3+\eta$.
(Note that all points of $S_B$ itself can be covered 
by including the single point $p_{-}$ at no cost.)
Choosing covering points in $S_B$ corresponds to assigning labels
in $B$ to vertices of $V$ in the Label Cover instance.
\item
Points in $S_L$ have infinite weight, so no points of
$S_L$ appear in the solution. Hence, either $p'_+$ or some
points of $S_A$ must be used to cover points of $S_L$. 
Specifically, a point in $S_L \in S_+$ incident on an included point
of $S_B \in S_-$ is at distance exactly $2+\eta$ from this point, 
and so it must be covered by some point of $S_A$ to which it
is connected, at distance 2 -- other points in $S_A$ are
farther than $2+\eta$.
Points of $S_L$ not incident on an included point of $S_B$
can be covered by $p'_+$, which at distance
$2+2\eta$ is still closer than any point in $S_B$.
(Note that all points in $S_A$ itself can be covered by including
a single arbitrary point of $S_A$, which at distance at most 1 is
closer than all other point sets.)
Choosing the covering point in $S_A$ corresponds to assigning labels
in $A$ to vertices of $U$ in the Label Cover instance, thereby inducing
a valid labeling for some edge and solving the Label Cover problem.
\end{enumerate}

As the cost of WNNC is determined only by the number of chosen points in
$S_A$, a solution of cost $c$ to WNNC is equivalent to choosing $c$
labels in $A$ in the Label Cover instance. It follows that it is NP-Hard
to approximate WNNC with weights $\{0,1,\infty\}$ to within a factor
$2^{\log^{1-o(1)}m}$.

\paragraph{Modifying weights.}
Before reducing WNNC to NNC, we note that the above 
reduction carries over to instances of WNNC with weights
in the set $\{1,m^2,m^4\}$:
Points $p_+,p_-,p'_+$ and all points in $S_B$ are assigned
weight 1 instead of 0. 
Points in $S_A$ are assigned weight $m^2$ instead of $1$.
And points in $S_E,S_L$ are assigned weight $m^4$ instead
of $\infty$.
Now, a trivial solution to this instance of WNNC is to take
all points of $S_A,S_B$ and the single point $p_+$:
The total cost of this solution is less than $m^3$, 
and this provides an upper bound on the optimal solution cost.
It follows that choosing any point of $S_E,S_L$ at cost
$m^4$, results in a solution that is greater than optimal
by a factor at least $m$. Further, choosing
all points of $S_B$ along with $p_+,p_-,p'_+$ amount to 
only an additive cost $m$. Hence, 
the cost of WNNC is asymptotically equal to the number of points 
of $S_A$ included in its solution.
It follows that WNNC with weights in the set 
$\{1,m^2,m^4\}$ is NP-hard to approximate within a factor of
$2^{\log^{1-o(1)}m}$.

\paragraph{Second reduction.}
We now reduce WNNC to NNC, and this requires that we mimic the
weight assignment of WNNC using the unweighted points of NNC.
We introduce the following gadget graph $G(w)$ which allows us to 
assign weight $w$ to any point:
Create a point set $T$ of size $w$ of contiguous points realizing
a $D$-dimensional $\ell_1$-grid of side-length $g' = w^{-1/D}$.
(Note that for $w > m$, $g' < g$.) 
Now replace each point $p \in T$ by twin positive and negative points
at mutual distance $\frac{g'}{2}$,
such that distance between a point replacing $p \in T$ to one replacing
any $q \in T$ is the same as the original distance from $p$ to $q$. 
$G(w)$ consists of $T$,
along with a single positive point at distance $10$ from all positive points
of $T$, and $10 + \frac{g'}{2}$ from all negative points of $T$,
and a single negative point at distance $10$ from all negative points
of $T$, and $10 + \frac{g'}{2}$ from all positive points of $T$.
By construction, the diameter of $G(w)$ is at most 1, while its scaled margin is 
$O(g')$.

Clearly, the optimal solution to NNC on $G(w)$ is to choose only the
two points not in $T$. If any point in $T$ is included in the solution,
then all of $T$ must be included in the solution: First the twin of the included point
must also be included in the solution. Then, any point at distance $g'$ from both twins
must be included as well, along with its own twin.
But then all points within distance $g'$ of the new twins must be
included, etc., until all points of $T$ are found in the solution.

Given an instance of NNC, we can assign weight 
$m^2$ or $m^4$ to a positive point $p$ by creating a gadget 
$G(m^2)$ or $G(m^4)$ for this point.
All points of the gadget are at distance
$10$ from $p$. If $p$ is not included in the 
NNC solution, then the cost of the gadget is only 2.
(Note that the distance from the gadget points to all other
points in the NNC instance is at least $10+g > 10 +\frac{g'}{2}$.)
But if $p$ is included in the NNC solution, then it is the nearest 
neighbor of the negative gadget points, and so all the gadget points must be included in 
the solution, incurring a cost of $m^2$ or $m^4$. A similar argument allows us to assign weight
to negative points of NNC. The scaled margin of the NNC instance is of size 
$O(g') = m^{-O(1/D)}$, which completes the proof of Theorem \ref{thm:hard}.

\section{Learning}
\label{sec:learn}

In this section, we apply Theorem~\ref{thm:alg}
to obtain improved 
generalization bounds
for binary classification in doubling metric spaces.
Working in the standard agnostic PAC setting,
we take the labeled sample $S$ to be drawn iid
from some unknown distribution over $\X\times\set{-1,1}$,
with respect to which all of our probabilities will be defined.

Our basic work-horse for proving generalization bounds
is the notion of a {\em sample compression scheme}
in the sense of \citet{DBLP:journals/ml/GraepelHS05},
where it is treated in full rigor.
Informally, a learning algorithm 
maps a sample $S$ of size $n$ to a hypothesis $h_S$.
It is a {$d$-sample compression scheme} if a sub-sample of size $d$
suffices to produce (and unambiguously determines)
a hypothesis that agrees with the labels
of all the $n$
points.
It is an 
{\em $\eps$-lossy}
$d$-sample compression scheme if a sub-sample of size $d$
suffices to produce a hypothesis that
disagrees with the labels of
at most $\eps n$ 
of
the $n$ sample points.

The algorithm 
need not know $d$ and $\eps$ in advance.
We say that 
the sample $S$
is {\em $(d,\eps)$-compressible} 
if 
the algorithm succeeds in finding
an
{$\eps$-lossy}
$d$-sample compression scheme for this particular sample.
In this case:

\begin{theorem}[\citet{DBLP:journals/ml/GraepelHS05}]
\label{thm:gen}
For any distribution over $\X\times\set{-1,1}$,
any $n\in\N$ and any $0<\delta<1$,
with probability at least $1-\delta$ over the random sample 
$S
$ of size $n$, the following holds:
\begin{enumerate}
\item[(i) ]If $S$ is 
{$(d,0)$-compressible}, then
$${\ds
\err(h_S) \le \oo{n-d}\paren{
(d+1)\log n
+\log\oo\delta}.
}$$
\item[(ii)] 
If $S$ is 
{$(d,\eps)$-compressible}, then
$${\ds
\err(h_S) \le
\frac{\eps n}{n-d}+\sqrt{
\frac{
(d+2)\log n
+\log\oo\delta}{2(n-d)}
}.
}$$
\end{enumerate}
\end{theorem}
The generalizing power of sample compression was independently 
discovered by \citet{warmuth86,MR1383093}, and later elaborated upon by 
\citet{DBLP:journals/ml/GraepelHS05}.
A ``fast rate'' version of Theorem~\ref{thm:gen}
was given in~\cite[Theorem 6]{gkn-jmlr17+aistats},
which provides
a smooth interpolation between the 
the
$(\log n)/n$ decay in the
lossless ($\eps=0$) regime to the 
$\sqrt{(\log n)/n}$ decay in the
lossy regime.

We now specialize the general sample compression result
of Theorem~\ref{thm:gen} to our setting.
In a slight abuse of notation, we will blur the distinction
between $S\subset\X$ as a collection of points in a metric space
and $S\in (\X\times\set{-1,1})^n$ as a sequence
of point-label pairs.
As mentioned in the preliminaries, there is no loss of generality in taking
$\diam(S)
=1.$
Partitioning the sample $S=S_+\cup S_-$
into its positively and negatively labeled subsets,
the {margin} induced by the sample is given by
$
\gamma(S) = 
\dist(S_+,S_-),
$
where
$
\dist(A,B):=
\min_{x\in A,x'\in B}\dist(x,x')$
for $A,B\subset\X.
$
Any $\tilde S\subseteq S$ induces
the nearest-neighbor classifier $h_{\tilde S}:\X\to\set{-1,1}$ via
\beqn
\label{eq:h_S}
h_{\tilde S}(x) = 
\begin{cases} 
+1 &\mbox{if } \dist(x,\tilde S_+)<\dist(x,\tilde S_-) \\ 
-1 & \mbox{else}. 
\end{cases} 
\eeqn

For $k\in\N$ and $\gamma>0$,
let us say that the sample $S$ is
$(k,\gamma)$-{\em separable}
if it admits a sub-sample
$S'\subset S$
such that
$|S\setminus S'|\le k$
and
$\marg(S')>\gamma$.
We observe,
as in \cite[Lemma 7]{gkn-aistats16},
that separability implies compressibility (the proof,
specialized to metric spaces, is provided for completeness):
\begin{lemma}
  \label{lem:sep-comp}
If $S$ is $(k,\gamma)$-separable then it is 
$
\paren{
\lceil 1/\gamma \rceil^{\ddim(S)+1}
,
{k}/{|S|}}$-compressible.
\end{lemma}
\bepf
Suppose $S'\subset S$ is a witness of 
$(k,\gamma)$-separability.
Being pessimistic, we will allow our lossy sample compression scheme
to mislabel all of $S\setminus S'$, but not any of $S'$, giving it a sample error
$\eps \le {k}/{|S|}$. Now by construction,
$S'$ is $(0,\gamma)$-separable, and thus a $\gamma$-net $\tilde S\subset S'$
suffices to recover the correct labels of $S'$ via $h_{\tilde S}$,
the $1$-nearest neighbor classifier induced by $\tilde S$
as in (\ref{eq:h_S}).
We bound the size of the $\gamma$-net
by
$
\lceil 1/\gamma \rceil^{\ddim(S)+1}
$
via (\ref{eq:packing}),
whence the compression bound.
\enpf

\begin{corollary}
\label{cor:gen}
With probability at least $1-\delta$,
the following holds:
If $S$ is 
$(k,\gamma)$-separable 
with witness $S'$
and $\tilde S\subseteq S'$ is a $\gamma$-net as in Lemma~\ref{lem:sep-comp},
then
\beq
\label{eq:Reg}
\err(h_{\tilde S}) &\le& 
\frac{k}{n-
\ell
}+\sqrt{
\frac{
(\ell+2)
\log n
+\log\oo\delta}{2(n-
\ell
)}
},
\eeq
where $\ell=\lceil 1/\gamma \rceil^{\ddim(S)+1}$.
\end{corollary}
{\bf Remark.}
It is instructive to compare the bound above to \cite[Corollary 2]{DBLP:journals/tit/GottliebKK14}.
Stated in the language of this paper, the latter upper-bounds
the $1$-NN generalization error in terms of the sample margin $\gamma$ and $\ddim(\X)$ by
\beqn
\label{eq:gkk10}
\eps+\sqrt{\frac2n\paren{d_\gamma\ln(34en/d_\gamma)\log_2(578n)+\ln(4/\delta)}},
\eeqn
where 
$
d_\gamma =\ceil{16/\gamma}^{\ddim(\X)+1}
$
and
$\eps$ is the fraction of the points in $S$ that violate the margin condition
(i.e., opposite-labeled point pairs less than $\gamma$ apart in $\dist$).
Hence, Corollary~\ref{cor:gen} is a considerable improvement over 
(\ref{eq:gkk10})
in at least three aspects. First, the data-dependent $\ddim(S)$ may be significantly smaller
than the dimension of the ambient space, $\ddim(\X)$.\footnote{
In general, $\ddim(S) \le c \ddim(\X)$ for some universal 
constant $c$, as shown in \cite{GK-13}.
} Secondly, the factor of
$16^{\ddim(\X)+1}$ is shaved off. Finally, 
(\ref{eq:gkk10}) relied on some fairly intricate fat-shattering arguments
\citep{alon97scalesensitive,299098}, while 
Corollary~\ref{cor:gen} is an almost immediate consequence of much simpler Occam-type results.

One limitation of Theorem~\ref{thm:alg} is that it 
requires the sample to be
$(0,\gamma)$-separable. The form of the bound in
Corollary~\ref{cor:gen}
suggests a natural Structural Risk Minimization (SRM) procedure:
minimize the right-hand size over $(\eps,\gamma)$.
A solution to this problem was (essentially) given in
\cite[Theorem 4]{DBLP:journals/tit/GottliebKK14}:
\begin{theorem}
Let $R(\eps,\gamma)$ denote the right-hand size 
of 
the inequality in 
Corollary~\ref{cor:gen}
and put
$
(\eps^*,\gamma^*) = \argmin_{\eps,\gamma} R(\eps,\gamma).
$
Then
\begin{enumerate}
\item[(i)] 
One may compute $(\eps^*,\gamma^*)$ in 
$O(n^{4.376})$ randomized time.
\item[(ii)] 
One may compute $(\tilde\eps,\tilde\gamma)$
satisfying
$
R(\tilde\eps,\tilde\gamma)\le 4R(\eps^*,\gamma^*)
$
in $O(\ddim(S)n^2\log n)$ deterministic time.
\end{enumerate}
Both solutions yield a witness $S'\subset S$
of $(\eps,\gamma$)-separability as a by-product.
\end{theorem}

Having thus computed the optimal (or near-optimal)
$\tilde\eps,\tilde\gamma$ with the corresponding sub-sample $S'$, 
we may now run the algorithm
furnished by Theorem~\ref{thm:alg}
on $S'$ 
and invoke the generalization bound in Corollary~\ref{cor:gen}.
The latter holds uniformly over all $\tilde\eps,\tilde\gamma$.
An algorithm closely related to the one outlined above
was recently shown to be Bayes-consistent
\cite{DBLP:conf/nips/KontorovichSW17}

\section{Experiments}
\label{sec:exper}
In this section we discuss experimental results. 
First, we will describe a simple heuristic
built upon our algorithm.
The theoretical guarantees in Theorem~\ref{thm:alg} feature a 
dependence on the scaled margin $\gamma$, and our heuristic
aims to give an improved solution in the problematic case where
$\gamma$ is small. Consider the following
procedure for obtaining a smaller consistent set.
We first extract a net $S_\gamma$ satisfying the guarantees of 
Theorem~\ref{thm:alg}. We then remove points from $S_\gamma$ 
using the following rule: 
for all $i\in\set{1,0, \ldots \lfloor \log \gamma \rfloor}$,
and for each $p \in S_\gamma$, if the distance from $p$
to all opposite labeled points in $S_\gamma$ is at least
$2 \cdot 2^i$, then retain $p$ in $S_\gamma$
and remove from $S_\gamma$ 
all other points strictly within distance 
$2^i-\gamma$ of $p$ (see Algorithm \ref{alg:heuristic}).
We can show that the resulting set is consistent:

\begin{lemma}
The above heuristic produces a consistent solution.
\end{lemma}
\bepf
Consider a point $p \in S_\gamma$, and assume without
loss of generality that $p$ is positive. If
$\dist(p,S^-_\gamma) \ge 2 \cdot 2^i$, then the 
positive net-points strictly within distance $2^i$ of $p$
are closer to $p$ than to any negative point in $S_\gamma$.
Now, some removed positive net-point $q$ may be the 
nearest neighbor for points of $S$ not in the net
(strictly within distance $\gamma$ of $q$), but these 
non-net points must be strictly within
distance $(2^i-\gamma) + \gamma = 2^i$ of $p$, and
so are closer to $p$ than to any negative net-point.
Note that $p$ cannot be removed
at a later stage in the algorithm, since its distance
from all remaining points is at least $2^i-\gamma$.
\enpf

\begin{algorithm}
\caption{Consistent pruning heuristic}
\label{alg:heuristic}
\begin{algorithmic}[1]
\State $S_\gamma$ is produced by Algorithm \ref{alg:bf} or \ref{alg:fast}
\ForAll {$i\in\set{1,0,\ldots,\lfloor \log \gamma \rfloor}$}
    \ForAll {$p \in S_\gamma$}
        \If {$p \in S^\pm_\gamma$ and $\dist(p,S^\mp_\gamma) \ge 2 \cdot 2^i$}
            \ForAll {$q \ne p \in S_\gamma$ with $\dist(p,q) < 2^i - \gamma$}                   
                \State $S_\gamma \leftarrow S_\gamma \backslash \{q\}$
            \EndFor
        \EndIf
    \EndFor
\EndFor
\end{algorithmic}
\end{algorithm}

\begin{figure*}
  \begin{center}
      \includegraphics[width=0.69\columnwidth]{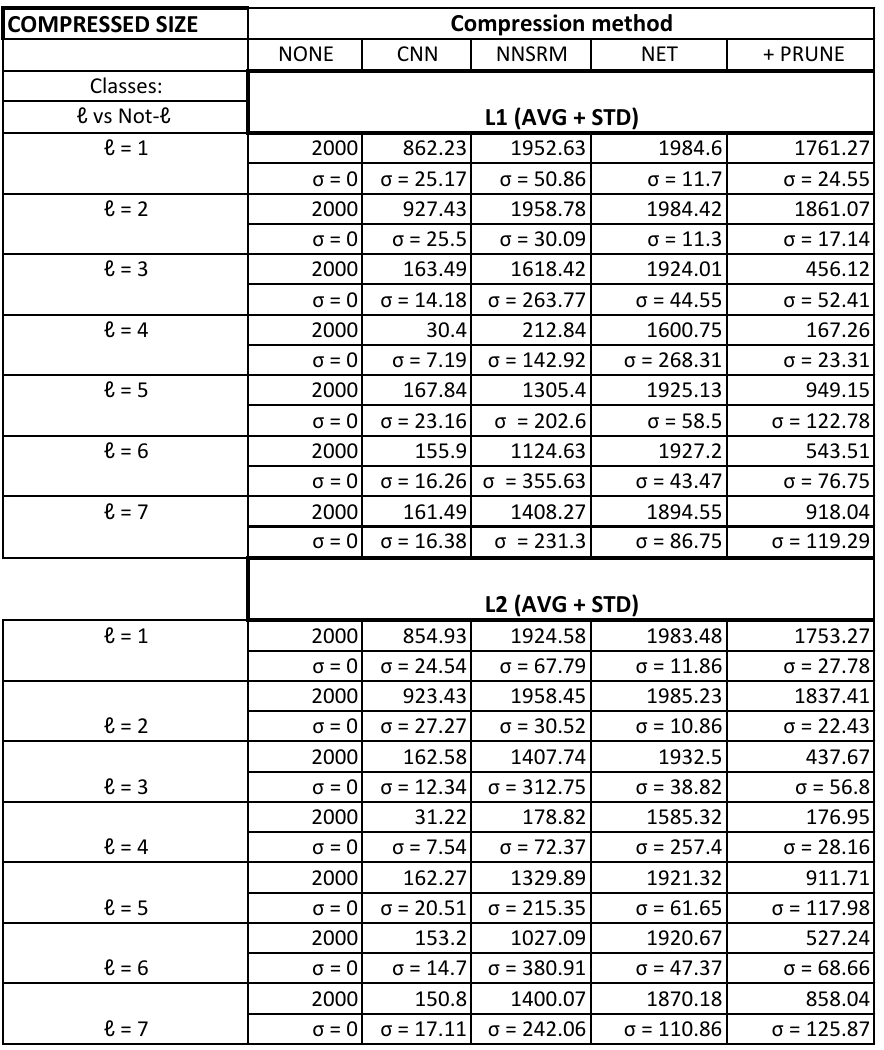}
    ${}$
    \includegraphics[width=0.69\columnwidth]{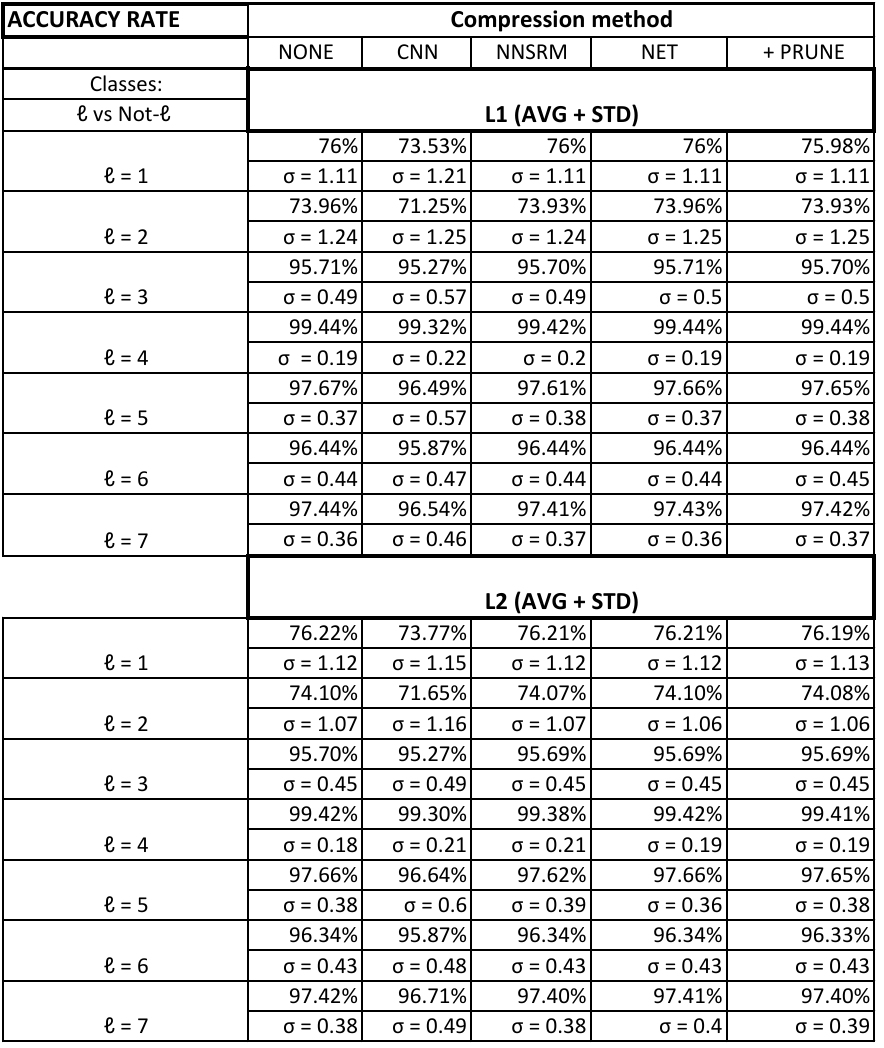}
    \caption{
      The amount of compression
and
generalization accuracy      
achieved the various methods.}
\label{fig:compr+acc}
\end{center}
\end{figure*}

As a proof of concept, we tested our sample compression algorithms
on several data sets from the UCI Machine Learning Repository,
involving  US Geological Survey data.\footnote{
\url{http://tinyurl.com/cover-data}
}
The data consisted of $7$ forrest cover types,
which we converted into $7$ binary classification problems
via the one-vs-all encoding. We note in passing
that the compression technique introduced here
is readily applicable in the multiclass setting
\cite{kon-weiss-2014}
(including a recent activised version
\cite{DBLP:journals/corr/KontorovichSU16-nips})
and even has a Bayes-consistent variant
\cite{DBLP:conf/nips/KontorovichSW17};
to maintain conceptual contiguity with the rest of the paper, we
only considered binary classification.
We ran
several
different nearest-neighbor condensing algorithms, as
well as the standard $1$-nearest neighbor as a baseline;
these are as follows:
\begin{itemize}
\item CNN --- Hart's original greedy rule,  \cite{DBLP:journals/tit/Hart68}
\item NNSRM --- Nearest Neighbor with Structural Risk Minimization,
  \cite{DBLP:journals/tnn/KaracaliK03}
\item NET --- the net-based approach prposed in this paper
\item +PRUNE --- net-based approach followed by pruning heuristic in Algorithm~\ref{alg:heuristic}.
\end{itemize}
Each classification task was performed with the $\ell_1$ and $\ell_2$
distances as the choice of metric, and the reported results are averaged
over $164$ trials.
Each average is accompanied by a standard deviation (denoted by $\sigma$).
In each trial,
the training set was constructed
by drawing
$1000$ positive examples
uniformly
at random from class $\ell$ and $1000$ negative examples
randomly
from the remaining classes (i.e., ``not $\ell$'');
a test set of size $2000$ was constructed analogously.
In Figure~\ref{fig:compr+acc}, we report the amount of compression
and
generalization accuracy
achieved by each of the methods.
Our algorithm compares favorably to NNSRM, achieving better compression
with similar generalization accuracy. NNSRM also has the much
slower runtime of $O(n^3)$. CNN achieved better compression, at the cost
of worse generalization accuracy.
One would expect that compression yields better generalization accuracy, 
but no improvements in accuracy were reflected in these experiments. 
This is an interesting avenue for future research.

\paragraph{Acknowledgement.}
We thank Michael Dinitz
and
Yevgeni Korsunsky
for helpful conversations.

{
\bibliographystyle{plain}
\bibliography{../../mybib}

\appendix

\section{Fast net construction}

In this section we provide an illustration of 
the fast net algorithm of Section \ref{sec:alg}.
For each point $p \in S$ we will record a single
covering point in each net $S_{2^i}$ -- this is
$P(p,i)$.
For each $p \in S_{2^i}$ we will maintain a list
$N(p,i)$ of neighbors in $S_{2^i}$, and also a list 
$C(p,i)$ of points in $S_{2^{i-1}}$ which are covered
by $p$. In the algorithm, we assume that these lists 
are initialized to null.

Although we have assumed there that the scaled margin
$\gamma$ is known a priori,
knowledge of $\gamma$ is not actually necessary:
We may terminate the algorithm when we encounter a net
$S_{2^i}$ where for all $p \in S_{2^i}$ and $q \in S$, if
$\dist(p,q) < 2^i$ then $p$ and $q$ are of the same label set.
Clearly, the net $i = \lfloor \log \gamma \rfloor$ 
satisfies this property 
(as may some other consistent net with larger $i$).
It is an easy matter to check the stopping condition
during the run of the algorithm, during the query for
$\dist(q,T)$.

\begin{algorithm}
\caption{Fast net construction}
\label{alg:fast}
\begin{algorithmic}[1]
\Require $S$
\State $p \leftarrow$ arbitrary point of $S$
\State $S_2 \leftarrow \{p\}$ 
\Comment {Top level contains a single point}
\ForAll {$q \in S$}
	\State $P(q,1) \leftarrow p$
	\Comment {$p$ covers all points}
\EndFor
\For {$i=1,0,\ldots,\lfloor \log \gamma \rfloor +1$}
    \ForAll {$p \in S_{2^i}$ and then $p \in S - S_{2^i}$}
        \State $T \leftarrow \cup_{r \in N(P(p,i),i)} C(r,i)$ 
                \Comment {Potential neighbors of $p$ in level $i-1$}
        \If {$\dist(p,T) < 2^{i-1}$}
            \State $P(p,i-1) \leftarrow$ point $r \in T$ with $\dist(p,r) < 2^{i-1}$
        \Else
            \State $S_{2^{i-1}} \leftarrow S_{2^{i-1}} \cup \{p\}$
            \Comment $p$ is placed in level $i-1$
            \State $C(P(p,i),i) \leftarrow C(P(p,i),i) \cup \{p\}$
            \Comment Update child list of $p$'s parent
            \ForAll {$r \in T$ with $\dist(p,r) < 4 \cdot 2^{i-1}$}
                \State $N(p,i-1) \leftarrow N(p,i-1) \cup \{r\}$
	        \Comment Build neighbor list for $p$	
                \State $N(r,i-1) \leftarrow N(r,i-1) \cup \{p\}$
	        \Comment Update $p$'s neighbors
            \EndFor
        \EndIf
    \EndFor
\EndFor
\end{algorithmic}
\end{algorithm}

\end{document}